\documentclass[10pt,conference]{IEEEtran}
\IEEEoverridecommandlockouts
\usepackage{cite}
\usepackage{amsmath,amssymb,amsfonts}
\usepackage{algorithmic}
\usepackage{graphicx}
\usepackage{textcomp}
\usepackage{booktabs}
\usepackage{siunitx}  % 用于数字对齐
\usepackage{enumitem}
\usepackage{ifsym}
\usepackage{adjustbox}
\usepackage{mathrsfs} % 额外花体字体包
\usepackage{multirow}
\usepackage[table,xcdraw]{xcolor}
\usepackage{inconsolata}
\usepackage[flushmargin]{footmisc}
\usepackage{makecell}
\usepackage{balance}
\usepackage{algorithm}

\def\BibTeX{{\rm B\kern-.05em{\sc i\kern-.025em b}\kern-.08em
    T\kern-.1667em\lower.7ex\hbox{E}\kern-.125emX}}
\begin{document}

\title{Evolvable Embodied Agent for Robotic Manipulation via Long Short-Term Reflection and Optimization}

\author{\IEEEauthorblockN{Jianzong Wang$^\dagger$, Botao Zhao$^\dagger$, Yayun He, Junqing Peng, Xulong Zhang$^\ast$\thanks{$^\dagger$Equal contribution.\newline$^\ast$Corresponding author: Xulong Zhang (zhangxulong@ieee.org)}}
\IEEEauthorblockA{\textit{Ping An Technology (Shenzhen) Co., Ltd.} 
Shenzhen, China}
}

\maketitle

\begin{abstract}
Achieving general-purpose robotics requires empowering robots to adapt and evolve based on their environment and feedback. Traditional methods face limitations such as extensive training requirements, difficulties in cross-task generalization, and lack of interpretability. Prompt learning offers new opportunities for self-evolving robots without extensive training, but simply reflecting on past experiences. However, extracting meaningful insights from task successes and failures remains a challenge. To this end, we propose the evolvable embodied agent (EEAgent) framework, which leverages large vision-language models (VLMs) for better environmental interpretation and policy planning. To enhance reflection on past experiences, we propose a long short-term reflective optimization (LSTRO) mechanism that dynamically refines prompts based on both past experiences and newly learned lessons, facilitating continuous self-evolution, thereby enhancing overall task success rates. Evaluations on six VIMA-Bench tasks reveal that our approach sets a new state-of-the-art, notably outperforming baselines in complex scenarios.
% Furthermore, the analysis of reflective optimization reveals high interpretability, offering deeper insights into the agent's self-evolution.
\end{abstract}


\section{Introduction}

Legacy robotic frameworks utilizing predetermined movements are broadly applied in logistics and manufacturing owing to their efficiency and uncomplicated design \cite{li2021overview, kyrarini2021survey}.
However, they still face challenges in large-scale data acquisition and the persistent sim-to-real gap \cite{zhao2020sim, kalashnikov2022scaling}.
The pursuit of versatile robotic agents has been significantly propelled by the recent breakthroughs in large language models (LLMs), introducing an innovative framework for the field \cite{achiam2023gpt, frieder2024mathematical, yuan2023well}.
Numerous studies have demonstrated that embedding Large Language Models within embodied systems substantially improves their adaptability across diverse tasks \cite{mu2024embodiedgpt, liang2023code}.
Additionally, vision language models (VLMs) have expanded their capabilities and been deployed in several practical applications \cite{Chen2024ALV,yang2025vlaser}.
However, (1) there is a lack of comprehensive research on large VLM-based embodied agents for enhancing generalizable robot manipulation, which could achieve robust and flexible environmental interpretation and policy planning.
(2) most of the current embodied agents do not possess the capacity to learn and adapt like humans, who enhance their task proficiency through accumulated experience.
We call this adaptive capacity \textbf{self-evolution}.

\begin{figure}[t]
	\centering
	\includegraphics[width=0.48\textwidth]{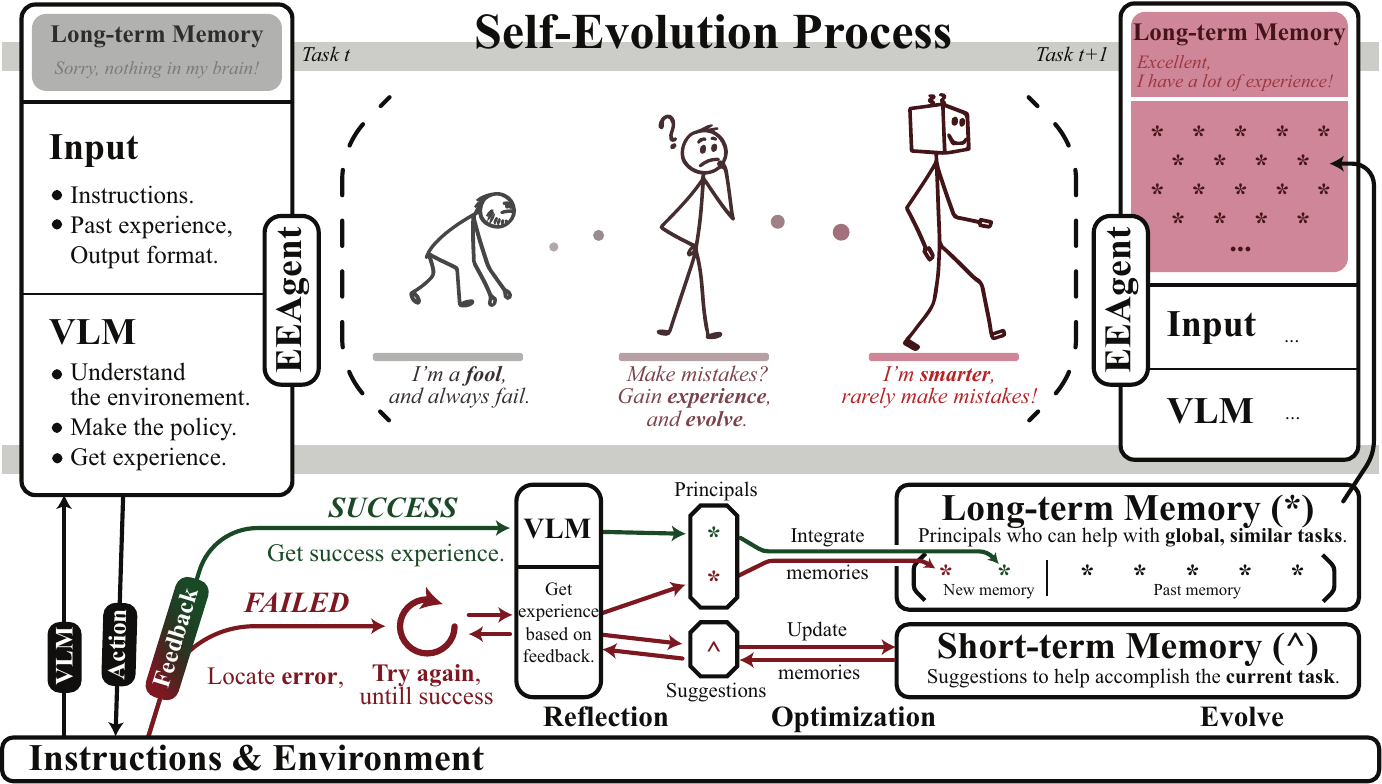}
	\caption{The motivation for introducing EEAgent is inspired by the human ability to learn from past successes and failures, extracting insights that can inform both present and future tasks.
		Building on this concept, we leverage short-term memory to guide current tasks and long-term memory to inform future tasks, enabling dynamic prompt optimization and fostering the self-evolution of the embodied agent.}
	\label{fig:motivation}
\end{figure}

There are several potential strategies to achieve embodied agents with self-evolution.
In conventional artificial intelligence (AI), incremental learning or model fine-tuning is commonly used to enhance task-specific performance \cite{wang2024comprehensive, chen2021decision}.
However, for general-purpose robotics, the sheer volume of model weights coupled with the heterogeneity of downstream applications make direct fine-tuning prohibitively expensive and challenging, while also posing issues of timeliness and interpretability.
Retrieval-Augmented Generation (RAG) \cite{lewis2020retrieval} has surfaced as a promising framework amidst the rapid proliferation of massive AI models \cite{zhu2024retrieval, majumderclin}, storing experiences in a memory bank for future task improvement.
Yet, this approach requires external storage and faces accuracy challenges in retrieval.
A novel technique enhancing model performance through reflection has recently gained attention.
This method, such as Reflexion \cite{shinn2024reflexion,  kim2023context}, generates language-based insights from past experiences to assist in task retries without parameter updates.
It incorporates a reflection process after task failures to identify causes and potentially increase future success rates.
Nevertheless, the ultimate success of this paradigm hinges significantly on the caliber and broad applicability of the derived reflections:
Not all reflected insights necessarily enhance generalization; 
Moreover, significant challenges remain in enabling VLMs to reflect accurately and provide truly generalizable insights.

Considering their inherent limitations, we propose a self-evolving embodied agent that integrates an optimized reflection process to enhance performance on similar tasks over time.
Additionally, we design an agent driven entirely by large VLMs to enable comprehensive environmental interpretation and policy planning.
To achieve this, we introduce an upstream environment interpreter that leverages VLMs to analyze complex environments, allowing the subsequent policy planner to generate more effective action sequences.

As shown in Fig.~\ref{fig:motivation}, inspired by the human learning process, particularly the role of memory in reflection and adaptation, our approach incorporates both long-term and short-term memories as interpretable prompts within a VLM.
Long-term memory accumulates generalizable knowledge for robust task execution, while short-term memory captures recent task-specific experiences to refine performance dynamically.
By continuously updating and integrating these memories, the agent progressively improves task success rates, achieving qualitative self-evolution with minimal trials.
To realize these objectives, we have made the following contributions:
\begin{itemize}[left=0pt]
	\item We propose an evolvable embodied agent (EEAgent) driven by VLMs, with an environment interpreter and policy planner to boost scalability and generalizability.
	\item We introduce a self-evolution strategy via long short-term reflective optimization (LSTRO), allowing iterative learning from past tasks to enhance performance and adaptability.
	\item Rigorous empirical testing on six distinct VIMA-Bench scenarios confirms that our technique yields state-of-the-art results, with the reflection-based evolution process exhibiting strong interpretability.
\end{itemize}

\begin{figure*}[t]
	\centering
	\includegraphics[width=0.98\textwidth]{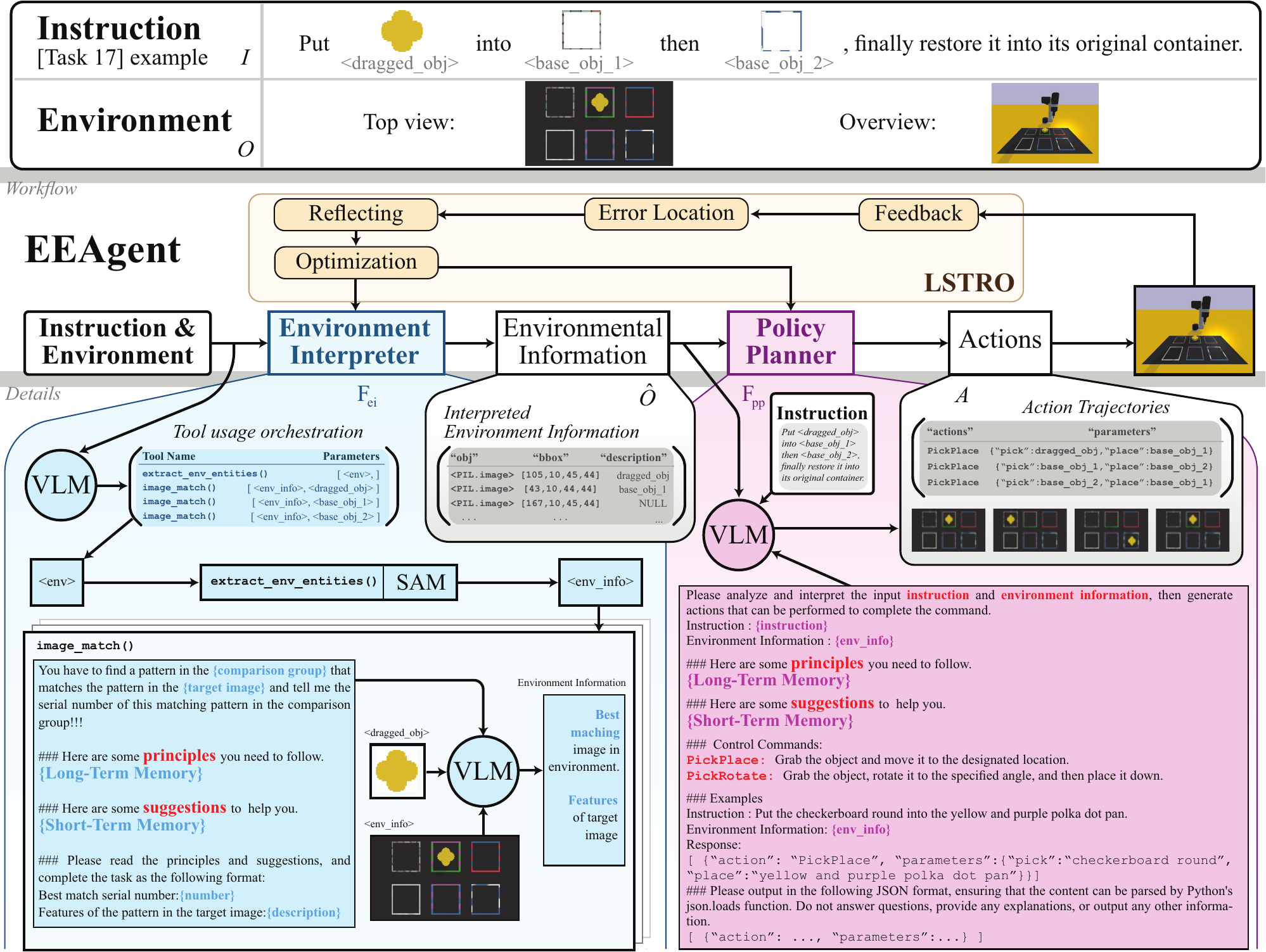}
	\caption{The overview architecture of EEAgent.
		The environment interpreter, is implemented using a function-calling VLM and predefined tools.
		It uses the SAM model to extract entities from the environment and a VLM to interpret them.
		The policy planner, implemented with an LLM, takes the instruction, environmental information, and prompt as input, producing an executable action sequence as output.}
	\label{fig:overview}
\end{figure*}

\section{Related Works}
\subsection{Robotic Manipulation with Pre-trained Models}
Traditional approaches to robotic manipulation have relied on predefined action sets and model-based control strategies, achieving efficiency in structured environments such as logistics and manufacturing~\cite{li2021overview,kyrarini2021survey}. These methods excel in simplicity and speed but encounter significant challenges in scaling to unstructured or dynamic settings. A major bottleneck of these approaches is their heavy reliance on collecting massive datasets for model optimization, the persistent sim-to-real gap that hinders deployment~\cite{zhao2020sim,kalashnikov2022scaling}, and poor cross-task generalization due to task-specific fine-tuning. For instance, reinforcement learning (RL)-based systems like those in~\cite{kalashnikov2022scaling} require extensive simulation trials to bridge domain discrepancies, yet they often fail in real-world variability, leading to brittle performance.

To address these issues, pre-trained foundation models have emerged as a paradigm shift. Multimodal foundation models, encompassing both VLMs and LLMs, facilitate zero-shot or few-shot adaptation by leveraging multimodal pre-training on vast datasets~\cite{achiam2023gpt,frieder2024mathematical,yuan2023well}. In robotics, this has facilitated generalizable manipulation through prompt-based planning. For example, Code as Policies (CaP)~\cite{liang2023code} uses LLMs to generate code snippets as executable policies, allowing flexible action synthesis from natural language instructions. Similarly, Instruct2Act~\cite{huang2023instruct2act} maps multimodal prompts directly to robotic actions, demonstrating improved generalization on benchmarks like VIMA-Bench~\cite{jiang2022vima,shao2025large}. However, these methods still suffer from hallucination in complex scenes—where VLMs misinterpret visual cues—and lack mechanisms for iterative improvement without retraining. Learning-based extensions, such as VIMA-20M~\cite{jiang2022vima} and VIMA-GPT~\cite{chen2021decision}, incorporate diffusion models or transformer-based sequence modeling for policy learning but demand prohibitive computational resources for fine-tuning, exacerbating the sim-to-real divide and reducing interpretability.

\subsection{Embodied Agents with Large Vision-Language Models}

Embodied agents integrate perception, reasoning, and action in physical environments, drawing inspiration from human-like interaction~\cite{mu2024embodiedgpt,Chen2024ALV}. Recent advances leverage VLMs for end-to-end environmental interpretation and policy planning, enabling scalable robotics without domain-specific engineering. EmbodiedGPT~\cite{mu2024embodiedgpt} employs chain-of-thought prompting with VLMs to simulate embodied reasoning, achieving cross-task transfer by aligning visual inputs with linguistic plans. Retrieval-augmented generation (RAG) variants, like those in~\cite{lewis2020retrieval}, augment agents with external memory banks of past trajectories, retrieving relevant examples to guide planning and mitigate forgetting in continual settings.

Despite these gains, current VLM-based agents exhibit gaps in robust visual grounding and adaptive planning. For instance, CLIN~\cite{majumderclin} uses continual learning to adapt LLMs rapidly but relies on external retrieval, introducing latency and retrieval inaccuracies in dynamic scenes. Moreover, while VLMs like GPT-4o~\cite{achiam2023gpt} and Gemini 1.5 Pro excel in zero-shot perception, they struggle with fine-grained entity extraction (e.g., distinguishing similar objects) without specialized tools like Segment Anything Model (SAM)~\cite{kirillov2023segment}. This leads to cascading errors in policy execution, particularly in combinatorial generalization tasks (e.g., novel object rearrangements) on benchmarks such as VIMA-Bench~\cite{jiang2022vima}. Solutions like Flamingo~\cite{alayrac2022flamingo} incorporate few-shot visual conditioning but remain static, unable to evolve from task feedback without parameter updates.

\subsection{Reflection and Memory Mechanisms for Self-Evolution}

Human-like self-evolution in agents requires mechanisms to reflect on experiences, consolidate knowledge, and adapt without exhaustive retraining~\cite{wang2024comprehensive}. Traditional continual learning~\cite{wang2024comprehensive} updates models incrementally but faces catastrophic forgetting in diverse robotic tasks. In the LLM era, reflection-based methods have gained traction as lightweight alternatives. Reflexion~\cite{shinn2024reflexion} introduces verbal reinforcement learning, where agents generate natural-language critiques of failures to refine future prompts, improving success rates in language tasks without gradients. Extensions like~\cite{kim2023context} add context-aware memory for instruction following, storing episodic traces to enhance environmental awareness.

To mimic human memory hierarchies, long short-term memory (LSTM)-inspired prompting separates generalized knowledge (long-term) from episodic details (short-term)~\cite{wangself}. Chain-of-Thought (CoT) prompting~\cite{wei2022chain} encourages sequential deductive processes to enhance model adaptability, while self-consistency (SC) variants~\cite{wangself} aggregate multiple reflections via voting or selection to reduce variance. Retrieval-augmented reflection~\cite{lewis2020retrieval,zhu2024retrieval,majumderclin} further integrates external databases, but retrieval noise limits reliability in real-time robotics.

These approaches address interpretability and cost but falter in extracting generalizable insights from sparse feedback (success/failure)~\cite{fang2025dualvla}. Reflections often yield task-specific anecdotes rather than transferable principles, and error localization remains ad-hoc without consistency checks between perception and actions~\cite{shinn2024reflexion}. Moreover, consolidating short-term experiences into long-term knowledge lacks dynamic optimization, leading to memory bloat or redundancy. Our work bridges these gaps by introducing LSTRO, a reflective optimization that dynamically refines VLM prompts with error-localized insights, enabling evolvable embodied systems demonstrating unprecedented adaptability on VIMA-Bench without external storage or fine-tuning.

\section{Methodology}
\subsection{Overview of Evolvable Embodied Agent}
The evolvable embodied agent (EEAgent) framework, is designed to enable a robot to comprehend its environment and orchestrate a series of physical operations in response to human directives and the environment acquired via the robotic optical module.
The EEAgent leverages a large VLM to facilitate visual reflection, crucial for high-level self-evolution.
The EEAgent, depicted in Figure \ref{fig:overview}, comprises two core modules:
(1) The environment interpreter, $\mathrm{F}_{\mathrm{ei}}$, is responsible for interpreting the relationship between the current environment and the provided instructions, as detailed in Section \ref{sec:env_interpreter}.
(2) The policy planner, $\mathrm{F}_{\mathrm{pp}}$, is tasked with generating an action sequence, $A$, based on the environmental information and instructions, as outlined in Section \ref{sec:policy_planner}.
To address the challenge of memory extraction, we introduce the long short-term reflection and optimization (LSTRO), as elaborated in Section \ref{sec:LSTRO}.

\subsection{Environment Interpreter}
\label{sec:env_interpreter}

Understanding the environment is a crucial step in robotic manipulation.
In this study, we leverage ChatGPT-4o~\cite{achiam2023gpt}, a sophisticated VLM, to implement the environment interpreter, $\mathrm{F}_{\mathrm{ei}}$.
As shown in Figure \ref{fig:overview}, we first utilize the function-calling capabilities of the LLM to generate sequences of function tools.
Specifically, we define four types of functional tools in this study:
(1) \texttt{extract\_env\_entities()}:
Implemented using the segment anything model (SAM)~\cite{kirillov2023segment}, extracts object entities from the current environment.
(2) \texttt{image\_match()}:
When a visual entity (image) is specified in the instructions, this function (with VLM) identifies the best-matched entities (most similar object) within the environment.
Shown in Figure \ref{fig:overview}, the prompt, will store memories and update strategy is detailed in Section \ref{sec:LSTRO}.
Also, the VLM also generates feature descriptions of the target image for self-evolution.
(3) \texttt{semantic\_match()}:
Used when the entities in the instructions are described textually, such as ``place the \textbf{checkerboard round} into the \textbf{yellow and purple polka dot pan}''.
It is implemented using a VLM with a similar prompt structure and working mechanism as \texttt{image\_match()}.
(4) \texttt{scene\_match()}:
This function handles many-to-many entity matching (e.g., "rearrange to this \textbf{scene}" task). Its implementation uses the two match tools iteratively to match each entity in the specified scene with corresponding ones in the current environment.

\subsection{Policy Planner}
\label{sec:policy_planner}

The policy planner, $\mathrm{F}_{\mathrm{pp}}$, is designed to generate a sequence of executable actions, $A$.
This module is implemented using an LLM, as illustrated in Figure \ref{fig:overview}.
Similar to the matching tools (e.g., \texttt{image\_match()}), the prompt for the policy planner consists of a task description, principles, suggestions, and an output format.
The update strategy is detailed in Section \ref{sec:LSTRO}.
However, unlike \texttt{image\_match()}, the policy planner utilizes a predefined action library based on the characteristics of the task, enabling the LLM to generate action sequences that fulfill the instructions using a limited set of available actions.
Since our experiments are conducted on VIMA-Bench \cite{jiang2022vima}, we can defined two actions for the action library: \texttt{PickPlace} and \texttt{PickRotate}.
Each action is accompanied by a description and parameter requirements.
For example, \texttt{PickPlace} is described as ``grab the object and move it to the designated location'' and requires the parameters ``pick\_location'' and ``place\_location''.
Additionally, we include examples to guide correct action sequence generation and constrain LLM to output sequences in JSON format. Using this policy planner, we effectively generate executable action sequences from environmental information and user instructions.

\subsection{Long Short-Term Reflection and Optimization}
\label{sec:LSTRO}

\subsubsection{Overall Strategy}

To alleviate the discrepancy between foundation models and physical robotic systems, it is essential to store memories that incorporates external knowledge.
Inspired by human cognition, which distinguishes between long-term and short-term memory, we propose a similar structure for the agent's knowledge.
Long-term memory (LM) stores generalized set of experiences that are applicable across a variety of tasks, while short-term memory (SM) holds task-specific information.
Over time, short-term memories may gradually consolidate into long-term memory, enhancing the agent's knowledge base.
Moreover, reflective processes—based on task successes and failures—allow the agent to update both memory systems with valuable insights.

To enhance the self-evolution capabilities of the task by dynamically updating the prompt (memories of past experiences described in natural language) within the EEAgent, we propose a long short-term reflection and optimization (LSTRO) approach.
As shown in Algorithm~\ref{alg:lstro}, both the environment interpreter and policy planner incorporate $\mathrm{LM}$ and $\mathrm{SM}$.
The $\mathrm{LM}$ is designed to generalize principles, or environments that exhibit global similarities.
Whereas, the $\mathrm{SM}$ contains specific suggestions only for the current instruction and environment.
In this study, we assume that feedback is either a success or failed, and they each have different strategies.
\textbf{Success:}
The reflective processes involve summarizing the successful experiences and integrate them into the $\mathrm{LM}$.
\textbf{Failed:}
The error is first located (described in \ref{sec:error_location}), and then both experiences of principles and suggestions from the failure are extracted to update both $\mathrm{LM}$ and $\mathrm{SM}$ concurrently.
The $\mathrm{LM}$ affect across tasks, whereas $\mathrm{SM}$ only affect the current task.
Through this process, specific $\mathrm{SM}$ experiences gradually consolidate into the $\mathrm{LM}$, ultimately transforming into general principles that improve overall performance.

\subsubsection{Error Location}
\label{sec:error_location}

Locating the source of errors is essential when a system task fails.
This study introduces two methods for detecting errors in the environment interpreter and the policy planner.

\noindent
\textbf{(a) Image-description consistency:}
Errors in the environment interpreter often stem from \texttt{image\_match()}, where the VLM may struggle to recognize fine-grained image features.
To address this, we propose an image-description consistency method.
Specifically, by supplementing test and reference images with additional descriptive features, we can guide the VLM to focus on finer details.
An LLM then evaluates the consistency of these descriptions.
If they match, the image matching process is considered successful; otherwise, an error is likely present in this step.
Essentially, this method integrates VLM-based perception with chain-of-thought reasoning to enhance error detection, leveraging additional image feature descriptions.

\noindent
\textbf{(b) Action-to-instruction consistency:}
For the policy planner, we propose an action-to-instruction consistency method.
Here, an LLM regenerates instructions from executed actions, which are then compared to the original instructions.
If the instructions match, the original action is likely correct.
If not, the action is likely erroneous.
Essentially, this approach leverages few-shot learning to align actions generated by the LLM, reducing potential errors.

\begin{algorithm}[t]
	\footnotesize
	\renewcommand{\algorithmicrequire}{\textbf{Input:}}
	\renewcommand{\algorithmicensure}{\textbf{Output:}}
	\caption{The process of LSTRO}
	\label{alg:lstro}
	\begin{algorithmic}[1]
		\REQUIRE  instructions $I$;
		environments $O$;
		environments interpreter $\mathrm{F}_{\mathrm{ei}}\left(\hat{O}|I, O\right)$;
		policy planner $\mathrm{F}_{\mathrm{pp}}(A|I, \hat{O})$;
		task number $N$.
		Define long-term memory as $\theta=\left\{\mathrm{LM}^{\mathrm{ei}}, \mathrm{LM}_{\mathrm{pp}}\right\}$,
		short-term memory as $\gamma=\left\{\mathrm{SM}_{\mathrm{ei}}, \mathrm{SM}_{\mathrm{pp}}\right\}$.
		\STATE Initialization: $\mathrm{LM}_{\mathrm{ei}}=\left[\ \right]$, $\mathrm{LM}_{\mathrm{pp}}=\left[\ \right]$;
		\FOR{$t=1$ \TO $N$}
		\STATE Initialization: $\mathrm{SM}_{\mathrm{pp}}=\text{NULL}$, $\mathrm{SM}_{\mathrm{ei}}=\text{NULL}$;
		\STATE \color{gray}{/* Environmental Interpretation \& Policy Planning */}\color{black}
		\STATE Obtain the environmental information $\hat{O}=\mathrm{F}_{\mathrm{ei}}(I, O)$;
		\STATE Obtain the action sequence $A=\mathrm{F}_{\mathrm{ei}}(I, \hat{O})$;
		\STATE Do actions, get $\text{Feedback}\in\left\{\text{SUCCESS}, \text{FAILED}\right\}$;
		\STATE \color{gray}{/* Long Short-Term Reflection and Optimization */}\color{black}
		\IF{$\text{Feedback}=\text{SUCCESS}$}
		\STATE \textbf{Reflection:} reflect LM principles of success with $p_{t}^{\mathrm{ei}}$, $p_{t}^{\mathrm{pp}}$;
		\STATE \textbf{Optimization:} integrate $p_{t}^{\mathrm{ei}}$, $p_{t}^{\mathrm{pp}}$ to long-term memory $\theta$;
		\ELSE
		\STATE Initialization: trials $k=0$;
		\WHILE{$\text{Feedback}\ne\text{SUCCESS}$ \OR $k<\text{MAX\_TRIALS}$}
		\STATE \textbf{Error location:} image-description consistency, and action-to-instruction consistency;
		\STATE \textbf{Reflection:} reflect LM principles of failure with $p_{t,k}^{\mathrm{ei}}$, $p_{t,k}^{\mathrm{pp}}$;
		\STATE \textbf{Reflection:} reflect SM suggestions of failure with $q_{k}^{\mathrm{ei}}$, $q_{k}^{\mathrm{pp}}$;
		\STATE \textbf{Optimization:} integrate long-term memory $\theta$ by $p_{t,k}^{\mathrm{ei}}$, $p_{t,k}^{\mathrm{pp}}$,
		and update short-term memory $\gamma$ by $q_{k}^{\mathrm{ei}}$, $q_{k}^{\mathrm{pp}}$.
		\STATE $k=k+1$;
		\ENDWHILE
		\ENDIF
		\ENDFOR
		\ENSURE Updated long-term memory $\theta$.
	\end{algorithmic}
\end{algorithm}

\begin{figure*}[t]
	\centering
	\includegraphics[width=0.98\textwidth]{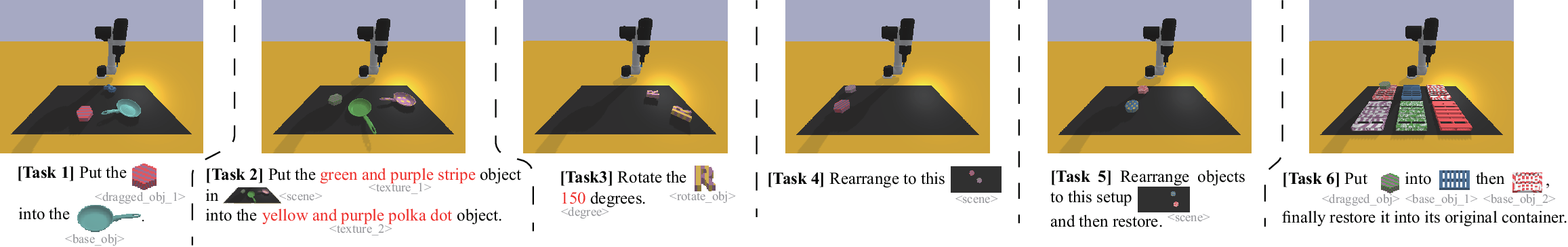}
	\caption{The instruction example of six VIMABench Task.}
	\label{fig6-tasks}
\end{figure*}

\begin{figure*}[t!]
	\centering
	\includegraphics[width=0.98\textwidth]{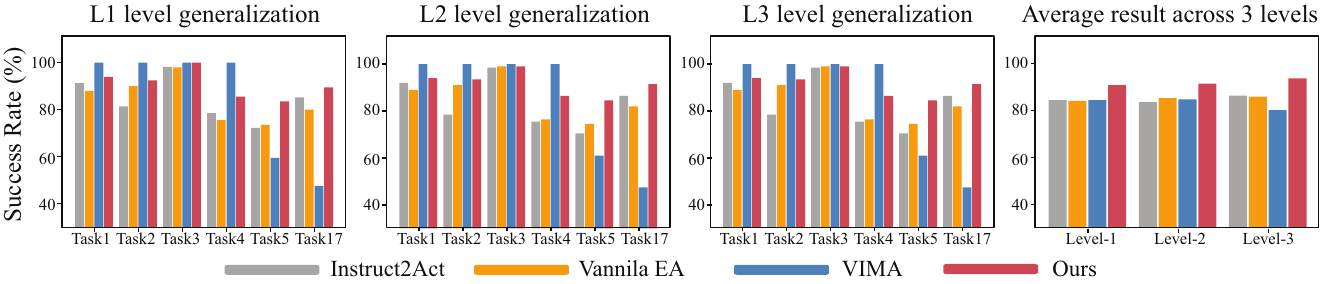}
	\caption{The performance comparison of Instruct2Act, Vanilla EA, VIMA, and our proposed method.}
	\label{fig:bars}
\end{figure*}

\subsubsection{Reflection}
Our EEAgent framework conduct the reflection based on the feedback of success and failed~\cite{li2025reflection}.
\textbf{For success cases}, a specific prompt will be used to guide the LLM in summarizing effective successful experiences, which are then stored in long-term memory to enhance future performance.
Considering the environment interpreter or policy planner may execute multiple times, all past dialogues will be concatenated, allowing the LLM to summarize the most simplified and useful aspects of the successful experiences.
\textbf{For failed cases}, the LLM recalls relevant memories, diagnoses the root causes of failure, and generates actionable suggestions to improve task execution.
Additionally, the LLM will produce concise and useful summaries of experiences to prevent recurring errors in future tasks.

\subsubsection{Optimization (memory updating)}
In this study, we propose an approach to evolving agents by dynamically updating their prompts based on insights gained from past experiences, which are summarized during a reflection process. Our method takes an additional step to address the inherent uncertainties in reflections generated by LLMs.
Reflections may introduce redundancies or contradictions with existing memories, and LLM outputs are prone to hallucinations, which can further increase uncertainty.
Over time, these issues may reduce the relevance of reflections for similar tasks in the future.
We first utilize the LLM to evaluate the generality of the reflective content.
Once the content passes this evaluation, it is combined with existing memory.
The LLM then consolidates semantically redundant information and resolves contradictions.
This process results in a refined long-term memory list that combines both new and existing knowledge.
Additionally, to manage the complexity of visual-language tasks, we limit the number of stored memories to a fixed maximum.

\section{Experiments and Results}

\subsection{Experimental Setup}

To validate our approach, we utilize the VIMA-Bench benchmark \cite{jiang2022vima}, which provides a diverse set of tasks that span both visual understanding and task planning.
Specifically, we selected six sub-tasks from VIMA-Bench for evaluation (illustrated in Figure \ref{fig6-tasks}).
They include:
(1) placing \{object1\} into \{object2\};
(2) placing an object with \{texture1\} in \{scene\} into an object with \{texture2\};
(3) rotating \{object1\} by \{angles\} degrees;
(4) rearranging objects to match a specified \{scene\};
(5) rearranging objects to a specific configuration and then restoring them to their original positions;
(6) placing \{object1\} into \{object2\}, then \{object3\}, and finally restoring them to their original containers.
These tasks align with the evaluation protocols used in Instruct2Act \cite{huang2023instruct2act}.
VIMA-Bench also provides a four-tier generalization evaluation protocol, which includes L1 (placement generalization), L2 (combinatorial generalization), L3 (novel object generalization), and L4 (novel task generalization).
Further details on the evaluation setup are available in \cite{jiang2022vima}.
Our system is built upon the ChatGPT-4o interface, utilizing the huge version of the SAM.
The number of max\_trails and maximum long memories are set to 3 and 20 respectively.
Notably, the system does not involve any local model training or inference.

\subsection{Baselines}
To thoroughly evaluate our approach, we conducted a comparison across three types of methods:
(1) \textbf{LLM-planning-based methods:}
In this category, we compared CaP \cite{liang2023code}, Instruct2Act \cite{huang2023instruct2act}, and CLIN \cite{majumderclin}.
While Instruct2Act can be directly applied to VIMA-Bench, CaP, and CLIN have been specific modifications under the Instruct2Act.
% which are detailed in the Appendix.
(2) \textbf{Learning-based methods:}
We compared several learning-based approaches, including VIMA-20M \cite{jiang2022vima}, VIMA-Gato \cite{reedgeneralist}, VIMA-Flamingo \cite{alayrac2022flamingo}, and VIMA-GPT \cite{chen2021decision}.
(3) \textbf{Prompting strategies:}
To assess the impact of our proposed LSTRO method on embodied agents, we evaluated various prompting strategies within the embodied agent framework introduced in this paper.
These strategies include:
(a) the vanilla embodied agent (Vanilla EA) without LSTRO,
(b) chain-of-thought prompting (CoT-Prompt) \cite{wei2022chain},
(c) ranking-based self-consistency (SC),
(d) selection-based SC,
(e) reflection-based SC \cite{wangself}.

\subsection{Main Performance}
\begin{table}[t]
	% \centering
	\caption{Overall performance comparison of VIMA-Bench in selected task generalizations.}
	\label{tab:overall}
    \adjustbox{max width=1\linewidth}{
	\setlength{\tabcolsep}{1.25pt}
	\renewcommand{\arraystretch}{1.0}
	\begin{tabular}{lrcccccc|c}
		\toprule
		\multicolumn{1}{l}{\multirow{1}{*}{Type}} & \multicolumn{1}{r}{\multirow{1}{*}{Methods}} & Task1 & Task2 & Task3 & Task4 & Task5 & Task6 & Avg.          \\
		\midrule
		\multirow{3}{*}{\makecell[l]{LLM                                                                                                                          \\Planning}}
		                                          & CaP                                          & 76.3  & 73.7  & 79.3  & 70.8  & 66.3  & 60.3   & 71.1          \\
		                                          & Instruct2Act                                 & 91.8  & 80.1  & 98.4  & 78.2  & 72.7  & 89.6   & 85.1          \\
		                                          & CLIN                                         & 81.7  & 77.7  & 80.7  & 72.8  & 70.3  & 72.7   & 76.0          \\
		\midrule
		\multirow{3}{*}{\makecell[l]{Learning                                                                                                                     \\-based}}
		                                          & VIMA-20M                                     & 99.3  & 100.0 & 100.0 & 99.5  & 58.7  & 41.5   & 83.2          \\
		                                          & VIMA-GPT                                     & 53.0  & 57.2  & 50.52 & 46.3  & 37.2  & 1.7    & 47.6          \\
		                                          & VIMA-Gato                                    & 50.7  & 56.2  & 38.7  & 52.3  & 36.0  & 1.0    & 45.9          \\
		                                          & VIMA-Flamingo                                & 52.8  & 55.0  & 54.5  & 51.0  & 40.3  & 0.5    & 49.2          \\
		% \addlinespace
		\midrule
		\multirow{5}{*}{\makecell[l]{Prompting                                                                                                                    \\Strategies}}
		                                          & Vannila EA                                   & 89.2  & 90.8  & 98.3  & 76.2  & 74.5  & 81.8   & 85.1          \\
		                                          & CoT-Prompt                                   & 91.7  & 92.8  & 99.2  & 82.5  & 80.0  & 88.3   & 88.2          \\
		                                          & S-C (Voting)                                 & 90.5  & 93.0  & 98.8  & 83.3  & 81.2  & 87.7   & 89.2          \\
		                                          & S-C (Select)                                 & 91.2  & 93.7  & 99.2  & 82.8  & 80.5  & 88.3   & 89.3          \\
		                                          & S-C (Reflect)                                & 90.8  & 92.5  & 99.0  & 82.7  & 79.8  & 87.2   & 88.7          \\
		% \addlinespace
		\midrule
		                                          & \textbf{EEAgent (Ours)}                      & 94.3  & 94.2  & 99.3  & 87.2  & 84.8  & 93.2   & \textbf{92.2} \\
		\bottomrule
	\end{tabular}
    }
\end{table}

As illustrated in Table \ref{tab:overall}, our results show a remarkable improvement over existing LLM-based methods across all six tasks.
In comparison to learning-based methods, our approach performs very close to VIMA-20M on tasks 1, 2, 3, and 4, while outperforming other methods in the comparison.
On more complex tasks, such as task 5 and task 6, our method has a clear advantage over training-based approaches, highlighting its ability to handle complicated tasks effectively.
In terms of prompting strategies, our LSTRO method consistently outperforms all existing strategies across all tasks, emphasizing its effectiveness in enhancing the environmental interpretation and task-planning capabilities of embodied agents.
Additionally, it is worth noting that even without LSTRO, the Vanilla EA also achieves a high success rate, particularly on more complex tasks.
This demonstrates that the embodied agent architecture we propose is inherently capable and robust, even without additional memory manipulations.

\begin{table}[t]
	% \centering
	\caption{Ablation studies were performed on different types of VLM with different SAM sizes.}
	\setlength{\tabcolsep}{5.5pt}
	\renewcommand{\arraystretch}{1.0}
    \adjustbox{max width=1\linewidth}{
	\begin{tabular}{lcccccc}
		\toprule
		\multicolumn{1}{l}{\multirow{1}{*}{Methods}} & Task1         & Task2         & Task3         & Task4         & Task5         & Task6        \\
		\midrule
		LLaVA-v1.5                                   & 80.5          & 79.7          & 84.2          & 75.2          & 71.5          & 69.2          \\
		Qwen2-VL                                     & 85.0          & 84.2          & 88.2          & 80.5          & 76.8          & 78.7          \\
		Gemini 1.5 Pro                               & 93.7          & 94.0          & 98.8          & \textbf{88.2} & \textbf{85.2} & 91.8          \\
		GPT-4o                                       & \textbf{94.3} & \textbf{94.2} & \textbf{99.3} & 87.2          & 84.8          & \textbf{93.2} \\
		\midrule
		Base-SAM                                     & 80.2          & 80.7          & 86.5          & 73.2          & 72.8          & 66.8          \\
		Large-SAM                                    & 88.3          & 89.5          & 96.2          & 84.7          & 82.2          & 87.5          \\
		Huge-SAM                                     & \textbf{94.3} & \textbf{94.2} & \textbf{99.3} & \textbf{87.2} & \textbf{84.8} & \textbf{93.2} \\
		\bottomrule
	\end{tabular}}
	\label{tab:ablation}
\end{table}

\subsection{Ablation Studies}
Table \ref{tab:ablation} details our assessment regarding the efficacy of our method using various LLMs, including LLaVA-v1.5 (7B), Qwen2-VL (70B), Gemini 1.5 Pro, and GPT-4o.
The results demonstrate that while smaller-scale models achieve satisfactory performance, larger and more powerful models lead to higher success rates.
Additionally, we tested segmentation models of different sizes, revealing that the base version significantly degrades performance.
This is due to the fact that accurate environment entity extraction plays a crucial role and is the basis for all subsequent operations.
Moreover, the segmentation model's quality directly impacts task execution.
Therefore, we recommend using the Large or Huge versions of SAM for more optimal results.

\begin{figure}[h]
	\centering
	\includegraphics[width=0.49\textwidth]{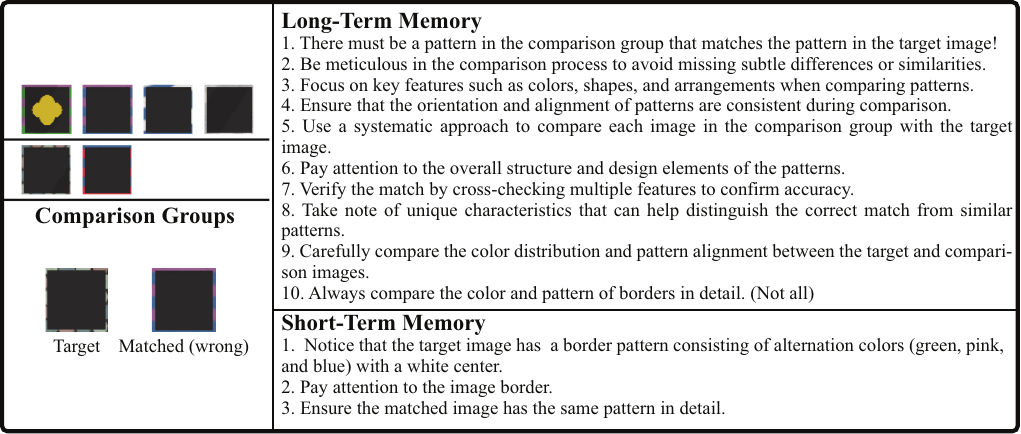}
	\caption{The illustration of partially learned long and short-term memory.}
	\label{fig:memory}
\end{figure}

\subsection{Analysis of Long Short-Term Memory}

We further analyzed the model's short- and long-term memory capabilities, using image entity matching as an example.
As illustrated in Figure \ref{fig:memory}, both memory types are highly interpretable.
Long-term memory significantly enhances image matching, much like the way humans leverage accumulated experience when mastering a skill.
In contrast, short-term memory is task-specific, acting as a reminder to address errors during task execution.
Interestingly, there is some overlap between short- and long-term memory, which can be attributed to their inherent similarities and the uncertainty in how long-term memory generalizes and infers information.
\section{Conclusions}
In this work, we proposed the Evolvable Embodied Intelligent Agent (EEAgent) framework, which integrates two core innovations:
(1) leveraging large vision-language models (VLMs) for environmental interpretation and policy planning,
(2) introducing a self-evolution strategy through the long short-term reflective optimization (LSTRO) mechanism, enabling the agent to learn iteratively from task execution and improve adaptability.
By utilizing tool-based operations rather than direct code generation, our framework allows dynamic updates to the prompts through self-reflection and error localization.
This facilitates continuous agent improvement or self-evolution.
We evaluated the framework on six sub-tasks from VIMA-Bench, demonstrating that our method outperforms existing LLM-based methods, exhibits superior generalization on complex tasks compared to training-based approaches, and provides significant advantages over other prompting strategies.
% This approach represents a key advancement in enabling embodied agents to self-update and evolve, which holds promise for the development of ultimate general-purpose robotics.
% We believe this framework holds promise for adapting a wider range of robots to perform more diverse tasks in the future.
\section{Acknowledgments}
This work was supported by the Shenzhen-Hong Kong Joint Funding Project (Category A) under Grant No. SGDX20240115103359001.

\bibliographystyle{IEEEtran.bst}
\bibliography{refs.bib}

\end{document}